\documentclass{article}

\usepackage{arxiv}

\usepackage[utf8]{inputenc} 
\usepackage[T1]{fontenc}    
\usepackage{hyperref}       
\usepackage{url}            
\usepackage{booktabs}       
\usepackage{amsfonts}       
\usepackage{nicefrac}       
\usepackage{microtype}      
\usepackage{lipsum}
\usepackage{epsfig} 
\usepackage{graphicx}
\usepackage{bm}
\usepackage{amsmath}

\newcommand{\argmax}{\arg\!\max}

\title{Reinforcement Learning and Control of a Lower Extremity Exoskeleton for Squat Assistance}

\author{
  Shuzhen Luo \\
  Department of Biomedical Engineering\\
  New Jersey Institute of Technology\\
  Newark, NJ 07102, USA\\
  \texttt{shuzhen.luo@njit.edu} \\
   \And
  Ghaith Androwis  \\
  Department of Biomedical Engineering\\
  New Jersey Institute of Technology\\
  Newark, NJ 07102, USA\\
  Kessler Foundation \\
  West Orange, NJ 07052, USA \\
  \texttt{ghaith.j.androwis@njit.edu} \\
   \And
  Sergei Adamovich \\
  Department of Biomedical Engineering\\
  New Jersey Institute of Technology\\
  Newark, NJ 07102, USA\\
  \texttt{sergei.adamovich@njit.edu} \\
   \And
  Hao Su \\
  Department of Mechanical Engineering\\
  The City University of New York\\
  City College, NY, 10023, USA\\
  \texttt{hao.su@ccny.cuny.edu} \\
   \And
  Xianlian Zhou\thanks{Corresponding author.}\\
  Department of Biomedical Engineering\\
  New Jersey Institute of Technology\\
  Newark, NJ 07102, USA\\
  \texttt{alexzhou@njit.edu} \\
}

\begin{document}
\maketitle

\begin{abstract}
A significant challenge for the control of a robotic lower extremity rehabilitation exoskeleton is to ensure stability and robustness during programmed tasks or motions, which is crucial for the safety of the mobility-impaired user. Due to various levels of the user's disability, the human-exoskeleton interaction forces and external perturbations are unpredictable and could vary substantially and cause conventional motion controllers to behave unreliably or the robot to fall down. In this work, we propose a new, reinforcement learning-based, motion controller for a lower extremity rehabilitation exoskeleton, aiming to perform collaborative squatting exercises with efficiency, stability, and strong robustness. Unlike most existing rehabilitation exoskeletons, our exoskeleton has ankle actuation on both sagittal and front planes and is equipped with multiple foot force sensors to estimate center of pressure (CoP), an important indicator of system balance. This proposed motion controller takes advantage of the CoP information by incorporating it in the state input of the control policy network and adding it to the reward during the learning to maintain a well balanced system state during motions. In addition, we use dynamics randomization and adversary force perturbations including large human interaction forces during the training to further improve control robustness. To evaluate the effectiveness of the learning controller, we conduct numerical experiments with different settings to demonstrate its remarkable ability on controlling the exoskeleton to repetitively perform well balanced and robust squatting motions under strong perturbations and realistic human interaction forces.
\end{abstract}


\section{Introduction}

Due to the aging population and other factors, an increasing number of people are suffering from neurological disorders, such as stroke, central nervous system disorder, and spinal cord injury (SCI) that affect the patient's mobility. Emerging from the field of robotics, robotic exoskeletons have become a promising solution to enable mobility impaired people to perform the activities of daily living (ADLs)~\cite{10.3389/fnbot.2018.00021, vouga2017twiice,8246978}. Lower-limb rehabilitation exoskeletons are wearable bionic devices that are equipped with powerful actuators to assist people to regain their lower leg function and mobility. With a built-in multi-sensor system, an exoskeleton can recognise the wearer’s motion intentions and assist the wearer’s motion accordingly~\cite{chen2016recent}.  
Compared to traditional physical therapy, rehabilitation exoskeleton robots have the advantages of providing more intensive patient repetitive training, better quantitative feedback, and improved life quality for patients~\cite{chen2016recent}. 

A degradation or loss of balance as a result of neuromusculoskeletal disorders or impairments is a common symptom, for instance, in patients with SCI or stroke. Balance training in the presence of external perturbations~\cite{horak1997postural} is considered as one of the more important factors in evaluating patients' rehabilitation performance. A rehabilitation exoskeleton can be employed for balance training to achieve static stability (quiet standing) or dynamic stability (squatting, sit-to-stand, and walking)~\cite{bayon2020can,mungai2020feedback, rajasekaran2015adaptive}. Squatting exercises are very common for resistance-training programs because their multiple-joint movements are a characteristic of most sports and daily living activities. In rehabilitation, squatting is commonly performed as an important exercise for patients during the recovery of various lower extremity injuries~\cite{SALEM2001424,crossley2011performance,yu2019design,mcginty2000biomechanical}. Squatting, which is symmetric by nature, can help coordinate bilateral muscle activities and strengthen weaker muscles on one side (e.g. among hemiplegia patients) or both sides. Compared to walking, squatting is often perceived to be safer for patients who are unable to perform these activities independently. In addition, the range of motion and the joint torques required for squatting are often greater than walking~\cite{yu2019design}. 
With a reliable lower extremity rehabilitation exoskeleton, performing squatting exercises without external help (e.g. from a clinician) will be a confidence boost for patients to use the exoskeleton independently. However, in order for the exoskeletons to cooperate with the human without causing risks of harm, advanced balance controllers to robustly perform squatting motion that can deal with a broad range of environment conditions and external perturbations need to be developed. 

Most existing lower extremity rehabilitation exoskeletons require the human operator to use canes for additional support or having a clinician or helper to provide balance assistance to avoid falling down. They often offer assistance via pre-planned trajectories of gait and provide limited control to perform diverse human motions. Some well known exoskeletons include the ReWalk (ReWalk Robotics), Ekso (Ekso bionics), Indego (Parker Hannifin), TWIICE~\cite{vouga2017twiice} and VariLeg~\cite{schrade2018development}. When holding the crutches, the patient's interactions with the environment is also limited~\cite{baud2019bio}. One example is that the patient is unlikely to perform squatting with long canes or crutches. 
Very few exoskeletons are able to assist human motions such as walking without the need of crutches or helpers with the exception of a few well known ones: the Rex (Rex Bionics)~\cite{Rex.org} and the Atalante (Wandercraft)~\cite{Atalante.org}. These exoskeletons free the user's hands, but come at the cost of an very low walking speed and an increased overall weight (38kg for the Rex, 60kg for the Atalante) and are very expensive~\cite{vouga2017twiice}. In this paper, we introduce a relatively light weight lower extremity exoskeleton that includes a sufficient number of degrees of freedom (DoF) with strong actuation. On each side, this skeleton system has a 1 DoF hip flexion/extension joint, a 1 DoF knee flexion/extension joint, and a 2-DoF ankle joint, which can perform ankle dorsi/plantar flexion as well as inversion/eversion that can swing the center of mass laterally in the frontal plane. Moreover, four force sensors are equipped on each foot for accurate measurement of ground reaction forces (GRFs) to estimate the center of pressure (CoP) so as to build automatic balance control without external crutches assistance. 

Designing a robust balance control policy for a lower extremity exoskeleton is particularly important and represents a crucial challenge due to the balance requirement and safety concerns for the user~\cite{chen2016recent,kumar2020learning}. First, the control policy needs to run in real-time with limited sensing and capabilities dictated by the exoskeletons. Second, due to various levels of patients' disability, the human-exoskeleton interaction forces are unpredictable and could vary substantially and cause conventional motion controllers to behave unreliably or the robot to fall down. Virtual testing of a controller with realistic human interactions in simulations is very challenging and the risk of testing on real humans is even greater. To the best of our knowledge, investigations presenting robust controllers against large and uncertain perturbation forces (e.g. due to human interactions) have rarely been carried out as biped balance control without perturbation itself is a challenging task. Most existing balance controller designs for such lower extremity rehabilitation exoskeletons focused mostly on the trajectory tracking method, conventional control like Proportional–Integral–Derivative (PID)~\cite{xiong2014research}, model-based predictive control~\cite{shi2019review}, fuzzy control~\cite{ayas2017fuzzy}, impedance control~\cite{hu2012training,karunakaran2020novel}, and momentum-based control for standing~\cite{bayon2020can}. Although the trajectory tracking approaches can be easily applied to regular motions, its robustness against unexpected large perturbations is not great. 
On the other hand, model-based predictive control could be ineffective or even unstable due to inaccurate dynamics modeling, and it typically requires a laborious task-specific parameters tuning. The momentum-based control strategies have also been applied to impose standing balancing on the exoskeleton~\cite{bayon2020can,8488066}, which was first applied in humanoid robotics to impose standing and walking balance~\cite{lee2012momentum, koolen2016design}. This method aimed to simultaneously regulate both the linear and angular component of the whole body momentum for balance maintenance with desired GRF and CoP at each support foot. The movement of system CoP is an important indicator of system balance ~\cite{lee2012momentum}. When the CoP leaves or is about to leave the support base, a possible toppling of a foot or loss of balance is imminent and the control goal is to bring the CoP back inside the support base to keep balance and stability. Although the CoP information can sometimes be estimated from robot dynamics~\cite{lee2012momentum}, the reliability of such estimation highly depends on the accuracy of the robot model and sensing of joint states. When a human user is involved, it is almost impossible to estimate the CoP accurately due to the difficulty to estimate the user's dynamic properties or real-time joint motions. Therefore, it is highly desired to obtain the foot CoP information directly and accurately. In most existing lower extremity rehabilitation exoskeletons, the mechanical structures of the foot are either relatively simple with no force or pressure sensors or with sensors but no capability to process the GRF and CoP information for real-time fall detection or balance control. In this work, a lower extremity rehabilitation exoskeleton with force sensors equipped on each foot for accurate estimation of CoP is presented. Inspired by the CoP-associated balance and stability~\cite{lee2012momentum}, this paper aims to explore a robust motion controller to encourage the system CoP to stay inside a stable region when subjected to the uncertainty of human interaction and perturbations. 

Recently, model-free control methods, like reinforcement learning (RL), promise to overcome the limitations of prior model-based approaches that require an accurate dynamic model. It has gained considerable attention in multi-legged robots control for their capability to produce controllers that can perform a wide range of complicated tasks~\cite{peng2016terrain,peng2017learning,2018-TOG-deepMimic,hwangbo2019learning,peng2020learning}. The RL-based balance control approach for lower extremity rehabilitation exoskeletons to perform squatting motion have not been investigated before, especially when balancing with a human strapped inside is considered. Since the coupling between the human and exoskeleton could leads to unexpected perturbation forces, it is highly desired to develop a robust controller to learn collaborative human-robot squatting skills. In this paper, we propose a novel robust control framework based on RL to train a robust control policy that operates on the exoskeleton in real-time so as to overcome the external perturbations and unpredictable varying human-exoskeleton force.  

The central contributions of this work are summarized in the following:
 \begin{itemize}
 	\item We build a novel RL-based motion control framework for a lower extremity rehabilitation exoskeleton to imitate realistic human squatting motion under random adversary perturbations or large uncertain human-exoskeleton interaction forces.
 	\item We take advantage of the foot CoP information by incorporating it into the state input of the control policy as well as the reward function to produce a balance controller that is robust against various perturbations.
 	\item We demonstrate that the lightweight exoskeleton can carry a human to perform robust and well-balanced squatting motions in a virtual environment with an integrated exoskeleton and full-body human musculoskeletal model.
 \end{itemize}

To demonstrate the effectiveness and robustness of the proposed control framework, a set of numerical experiments under external random perturbations and varying human-exoskeleton interactions are conducted. Dynamics randomization is incorporated into the training to minimize the effects of model inaccuracy and prepare for sim-to-real transfer. 

\section{Exoskeleton and Interaction Modeling}

\subsection{Mechanical Design of a Lower Extremity Robotic Exoskeleton}

\begin{figure}[h!]
\begin{center}
\includegraphics[width=0.9\columnwidth]{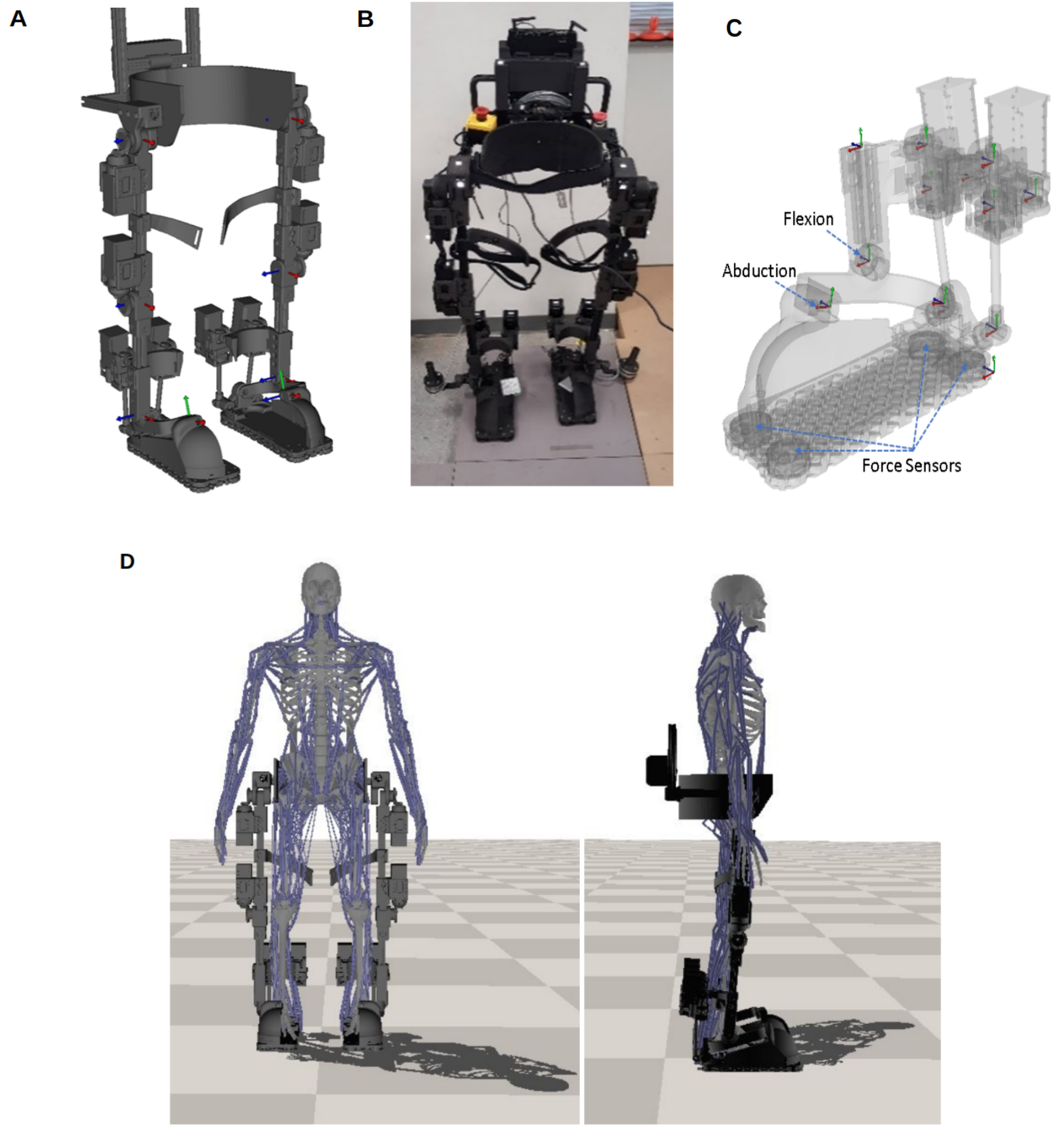}
\end{center}
\caption{The lower extremity exoskeleton and integrated human-exoskeleton model. \textbf{(A)} the physical prototype of the exoskeleton. \textbf{(B)} its multibody dynamics model. \textbf{(C)} the foot model with force sensors locations and the two independent ankle DoFs indicated. The joint axes of the model are displayed as well. \textbf{(D)} Integrated human and lower extremity exoskeleton model.}\label{fig:exo_device}
\end{figure}

A lower extremity robotic exoskeleton device~\cite{androwis2017research} is currently under development by the authors to assist patients with ADL, such as balance, ambulation and gait rehabilitation. In  Figure~\ref{fig:exo_device}A, the physical prototype of this exoskeleton is shown. The total mass of the exoskeleton is $20.4kg$ and the frame of the exoskeleton has been manufactured with Onyx (Markforged’s nylon with chopped fiber) reinforced by continuous carbon fiber between layers, using Markforged’s Mark Two printer (Markforged, INC, MA). The exoskeleton has 14 independent DoFs (including 6 global DoFs for the pelvis root joint): 1 DoF for the hip flexion/extension joint, 1 DoF for the knee flexion/extension joint, and 2 DoFs for the ankle dorsiflexion/plantarflexion and inversion/eversion joint on each side of the body. The 6-DoF pelvis (root) joint is a free joint and unactuated, and the rest 8 DoFs on both sides are actuated. Unlike most commercial rehabilitation exoskeletons that have either passive or fixed ankles, the ankle of our system includes powered 2-DoF to assist with dorsiflexion/plantarflexion and inversion/eversion~\cite{nunez20172}. All joints of the robotic exoskeleton are powered by smart actuators (Dynamixel Pro Motor H54-200-S500-R). 

Both hip and knee joints are driven by bevel gears for compact design. The ankle is actuated by two parallel motors that are attached to the posterior side the shank support and operate simultaneously in a closed-loop to flex or abduct the ankle. Figure~\ref{fig:exo_device}C shows the foot model of our exoskeleton system. At the bottom of each foot plate, four 2000N 3-axis force transducers (OptoForce Kft, Hungary) are installed to measure GRFs. These measured forces can be used to determine the CoP in real-time. The CoP is the point on the ground where the tipping moment acting on the foot equals zero~\cite{sardain2004forces}, with the tipping moment defined as the moment component that is tangential to the foot support surface. Let $\bm{O}_i$ denote the location of the $i-th$ force sensor and $\bm{F}_i$ be its measured force, $\bm{n}$ denote the unit normal vector of the foot support surface, and $\bm{C}$ denote the CoP point where the tipping moment vanishes: 
\begin{equation}
  \left[\sum_{i}^{4} (\bm{O}_i\bm{C} \times \bm{F}_i)\right] \times \bm{n} = \bm{0}
\label{CoP_computation}
\end{equation}
from which $\bm{C}$ can be computed.

\subsection{Exoskeleton Modeling}
A multibody dynamics model of the exoskeleton, shown in Figure~\ref{fig:exo_device}, is created with mass and inertia properties estimated from each part’s 3D geometry and density or measured mass. Simple revolute joints are used for the hips and knees. Each ankle has 2 independent rotational DoFs but their rotation axes are located at different positions (Figure~\ref{fig:exo_device}C). These 2 DoFs are physically driven by the closed-loop of two ankle motors together with linkage of universal joints and screw joints. When extending or flexing the ankle, both motors work in sync to move the two long screws up or down simultaneously. When adducting or abducting the ankle, the motors move the screws in opposite directions. The motors can generate up to 160$Nm$ dorsi/plantar flexion torque~\cite{nunez20172}. In this study, we do not directly control these ankle motors. Instead, we assume the two ankle DoFs can be independently controlled with sufficient torques from these motors.

\subsection{Modeling of Human Exoskeleton Interactions}
\label{human_interaction}
To simulate realistic human exoskeleton interactions, the exoskeleton is integrated with a full body human musculoskeletal model~\cite{lee2019scalable} that has a total mass of $61kg$ and includes 284 musculotendon units. The assembled exoskeleton has straps around the hip, femur and tibia to constraint the human motion. In our model, the root of the human musculoskeletal model (pelvis) is attached to the exoskeleton hip through a prismatic joint that allows relative movement only along the vertical (up \& down) direction (all rotations and the translations along the lateral and fore-and-aft directions are fixed). Meanwhile, an elastic spring with stiffness $k_h = 10000N/m$ is utilized to generate a vertical force and simulate the interaction between the human pelvis and the robot waist structure. Similarly, we use four springs (arranged in 90 degrees apart) with stiffness $k_f = 2000N/m$ to simulate the connections between the human femur and the strap on the robot femur. For the human tibia, four springs with stiffness $k_t = 2000N/m$ are also used to connect with the robot tibia.  
During motion, spring forces are generated and applied on both human and exoskeleton. The joints at the upper limb of the human are assumed as weld joints. We also assume there is no relative motion between the human foot and the exoskeleton foot due to tight coupling and model that as a welding constraint. In Figure~\ref{fig:exo_device}D, the integrated human musculoskeletal and exoskeleton model is shown. 


\section{Learning Controller for Balanced Motion}

\begin{figure}[h!]
\begin{center}
\includegraphics[width=1\columnwidth]{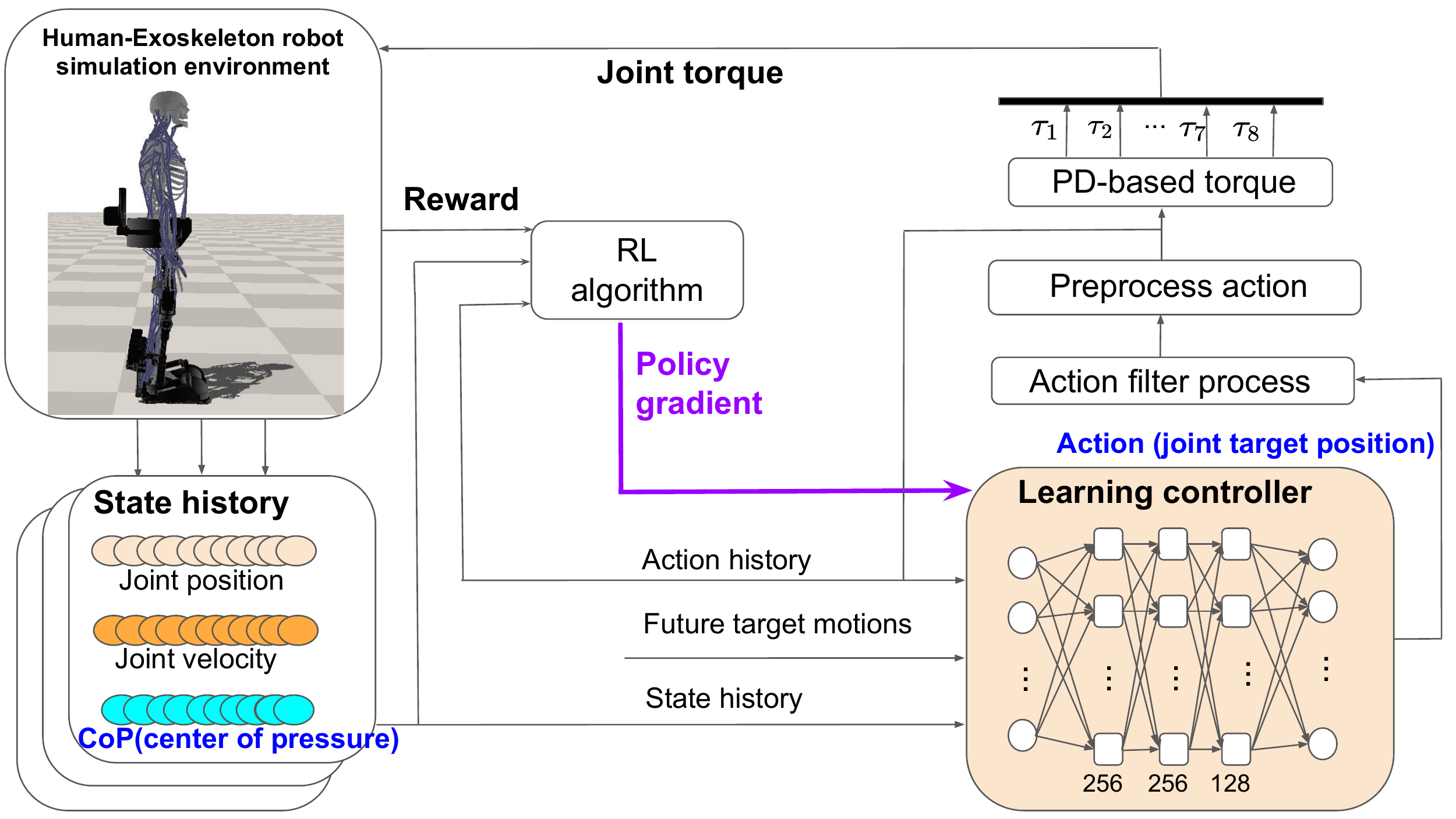}
\end{center}
\caption{The overall learning process of the integrated robust motion controller. We construct the learning controller as a Multi-Layer Perception (MLP) neural network that consists of three hidden layers. The neural network parameters are updated using the policy gradient method. The network produces joint target positions, which are translated into torque-level commands by PD (Proportional Derivative) control. }
\label{fig:StateDiagram}
\end{figure}

In this section, we propose a robust motion control training and testing framework based on RL that enables the exoskeleton to learn squatting skill with strong balance and robustness. Figure~\ref{fig:StateDiagram} shows the overall learning process of the motion controller. The details of the motion controller learning process will be introduced in the following sections. 

\subsection{Reinforcement Learning with Motion Imitation} 

The controller (or control policy) is learned through a continuous RL process. We design the control policy through a neural network with parameters $\theta$, denoting the weights and bias in the neural network. The control policy can be expressed as $\pi_\theta(a|s)$ and the agent (neural network model) learns to update the parameters $\theta$ to achieve the maximum reward. The learning controller (i.e. control policy network) is implemented as a Multi-Layer Perception (MLP) network that consists of three fully connected layers and ReLU as the activation function, as illustrated in Figure~\ref{fig:StateDiagram}, the sizes of three layers are set to 256, 256 and 128, respectively. At every time step $t$, the agent observes the state $s_t$ from the environment, and selects an action $a_t$ according to its control policy \(\pi(a_t|s_t)\). \(\pi(a_t|s_t)\) is in the form of the probability distribution of actions in a given state. The action $a_t$ is then converted to control commands that drive the exoskeleton in the environment, which results a new state $s_{t+1}$ and a scalar reward $r_t$ immediately. The objective is to learn a control policy that maximizes the discounted sum of reward:
\begin{equation}
\pi^*= \argmax_{\pi} \mathbb{E}_{\tau \sim p(\tau|\pi)}[\ \sum_{t=0}^{T-1}\gamma^tr_t]\
\label{reward}
\end{equation}
where $\gamma \in (0,1)$ is the discount factor,  $\tau$ is the trajectory $\{(s_0,a_0,r_0),(s_1,a_1,r_1), \cdots\}$ and $p(\tau|\pi)$ denotes the likelihood of a trajectory $\tau$ under a given control policy $\pi$, and $T$ is the horizon of an episode. 

The control policy is task- or motion-specific and starts with motion imitation. Within the learning process (Figure~\ref{fig:StateDiagram}), the input of the control policy network is defined by $s=\{p_{t-2:t},v_{t-2:t},c_{t-2:t},a_{t-3:t-1},\hat p_{t+1:t+6}\}$, in which $p$ and $v$ are joint positions and velocities of the exoskeleton, 
and $a_{t-3:t-1}$ represents the action history of three previous steps. To learn a particular skill, we utilize the corresponding target joint pose from the task motion at six future time-steps $\hat p_{t+1:t+6}$ as the motion prior for feasible control strategies. Considering the importance of CoP as an indicator of system balance and its ready availability from our exoskeleton design, we incorporate the CoP position history $c_{t-2:t}$ into the state as a feedback to the controller. As summarized in Figure~\ref{fig:StateDiagram}, the learning controller is given as input the combination of the state history, the action history and the future target motions and outputs the joint target positions as the actions. The use of task motion data alleviates the need to design task-specific reward functions and thereby facilitates a general framework to learn a diverse array of behaviors.

\subsection{Learning with Proximal Policy Optimization (PPO)} 

To train the control policy network, we use a model-free RL algorithm known as Proximal Policy Optimization (PPO). An effective solution to many RL problems is the family of policy gradient algorithms, in which the gradient of the expected return with respect to the policy parameters is computed and used to update the parameters $\theta$ through gradient ascent. 
PPO is a model-free policy gradient algorithm that samples data through interaction with the environment and optimizes a “surrogate” objective function~\cite{schulman2017proximal}. It utilizes a trust region constraint to force the control policy update to ensure that the new policy is not too far away from the old policy. The probability ratio $r_t(\theta)$ is defined by: 
\begin{equation}
r_t(\theta)= \frac{\pi_\theta(a_t|s_t)}{\pi_{\theta_{old}}(a_t|s_t)}.
\label{probability_ratio}
\end{equation} 
This probability ratio is a measure of how different the current policy is from the old policy $\pi_{\theta_{old}}$ (the policy before the last update). A large value of this ratio means that there is a large change in the updated policy compared to the old one. PPO also introduces a modified objective function that adopts clipped probability ratios which forms a pessimistic estimate of the policy's performance and avoids a reduction in performance during the training process. The "surrogate" objective function is described by considering the clipped objective:
\begin{equation}
L(\theta)= \mathbb{E}_t \left[min \left(r_t(\theta)\hat A_t, clip(r_t(\theta), 1-\epsilon, 1+\epsilon)\hat A_t \right)\right]
\label{surrogate_objective_function}
\end{equation}
where $\epsilon$ is a small positive constant which constrain the probability ratio $r_t(\theta)$. $\hat A_t$ denotes the advantage value at time step $t$. $clip(\cdot)$ is the clipping function. Clipping the probability ratio discourages the policy from changing too much and taking the minimum results in using the lower, pessimistic bound of the unclipped objective. Thus any change in the probability ratio is included when it makes the objective worse, and otherwise is ignored. This can prevent the policy from changing too quickly and leads to more stable learning. The control policy can be updated by maximizing the clipped discounted total reward in Equation~(\ref{surrogate_objective_function}) with a gradient ascent.

\subsection{Proportional-Derivative(PD)-based torque} 

It has been reported that the performance of RL for continuous control depends on the choice of action types~\cite{peng2017learning}. It works better when the control policy outputs PD position targets rather than joint torque directly. In this study, we also let the learning controller output the joint target positions as actions (shown in Figure~\ref{fig:StateDiagram}). 
To obtain smooth motions, actions from the control policy network are first processed by a second low-pass filter before being applied to the robot. Our learning process allows the control policy network and the environment to operate at different frequencies since the environment often requires a small time step for integration. 
During each time integration step, we apply preprocessed actions that are linear interpolated from two consecutive filtered actions. Then the preprocessed actions are specified as PD targets and the final PD-based torques applied to each joint are calculated as
\begin{equation}
\tau = clip\left( k_p(a_t-p_t)-k_v\dot p_t, - \hat \tau, \hat \tau \right)
\label{PD control}
\end{equation}
where $k_p$ and $k_v$ are proportional gain and differential gain respectively.
The function $clip(\cdot)$ returns the upper bound $\hat \tau$ or the lower bound$-\hat\tau$ if the torque $\tau$ exceeds the limit. 

\subsection{Reward Function}
We design the reward function to encourage the control policy to imitate a target joint motion $\hat p_t$ of the exoskeleton while maintaining balance with robustness. The reward function consists of pose reward $r_t^p$, velocity reward $r_t^v$, end-effector reward $r_t^e$, root reward $r_t^{root}$, center of mass reward $r_t^{com}$, CoP reward $r_t^{cop}$, and torque reward $r_t^{torque}$, which is defined by:
\begin{equation}
r_t = w^pr_t^p+ w^v r_t^v+ w^er_t^e+ w^{root}r_t^{root}+ w^{com}r_t^{com} + w^{cop}r_t^{cop}+ w^{torque}r_t^{torque}
\label{rewardfunction}
\end{equation}
where $w^p,w^v,w^e,w^{root},w^{com},w^{cop}, w^{torque}$ are their respective weights. The pose reward $r_t^p$ and velocity reward $r_t^v$ match the current and task (target) motions in terms of the joint positions $p_t$ and velocities $\dot p_t$:
\begin{equation}
r_t^p = \exp[-\sigma_p \sum \limits_j||\hat p_t^j-p_t^j||^2]
\label{pos_reward}
\end{equation}
\begin{equation}
r_t^v = \exp[-\sigma_v \sum \limits_j||\hat{\dot p}_t^j-\dot p_t^j||^2]
\end{equation}
where $j$ is the index of joints, $\hat p_t^j$ and $\hat{\dot p}_t^j$ are target position and velocity respectively. The end-effector reward is defined to encourage the robot to track the target positions of selected end-effectors:
\begin{equation}
r_t^e=\exp[-\sigma_e\sum \limits_i || \hat x_t^i-x_t^i||^2]
\end{equation}
where $i$ is the index of the end-effector. Let $x_t^i$ be the position of end-effectors, that include left foot and right foot, relative to the moving coordinate frame attached to the root (waist structure). The end-effectors motions are supposed to match well if the joint angles match well, and vice versa. 

We also design the root reward function $r_t^{root}$ to track the task root motion including the root's position $\hat x_t^{root}$ and rotation $\hat q_t^{root}$. 
\begin{equation}
r_t^{root} = \exp[-\sigma_{r1}||\hat x_t^{root}- x_t^{root}||^2-\sigma_{r2}||\hat q_t^{root}-q_t^{root}||^2]
\end{equation}

The overall center of mass reward $r_t^{com}$ is also considered in the learning process, enabling the control policy to track the target height during the complete squatting cycle. 
\begin{equation}
r_t^{com} = \exp[-\sigma_{com}||\hat x_t^{com}- x_t^{com}||^2]
\end{equation}

Besides the use of CoP positions in the state input as a feedback from the controller, we also introduce a CoP reward function to encourage the current CoP position $c_t^{cop}$ to stay inside a stable region around the center of the foot support. By considering the geometric of the foot in the lower extremity exoskeleton design, the stable region for foot CoP is defined as a rectangle area $S$ around the foot geometric center whose width and length are set to $7cm$ and $11cm$ respectively (narrower in the lateral direction than forward direction). And the CoP reward function is expressed as
\begin{equation}
r_t^{cop}=\left\{
\begin{aligned}
	& \exp[-\sigma_{cop}||D(c_t^{cop}, S)||^2], \text{if~} c_t^{cop} \in S\\
	& 0,  \ \text{if~} c_t^{cop}\notin S\\
\end{aligned}
\right.
\label{cop_reward}
\end{equation}
where $D(\cdot, \cdot)$ is the Euclidean distance between CoP and the center of $S$. The goal of this CoP reward is to encourage the controller to predict an action that will improve the balance and robustness of the exoskeleton's motion.

At last, we design the torque reward to reduce energy consumption or improve efficiency and to prevent damaging joint actuators during the deployment. 
\begin{equation}
r_t^{torque} = \exp[-\sigma_{torque}\sum_i||\tau_i||^2]
\label{torque_reward}
\end{equation}
where $i$ is the index of joints.

\subsection{Dynamics Randomization}
Due to the model discrepancy between the physics simulation and the real-world environment, well-known as reality or sim-to-real gap~\cite{yu2018policy}, the trained control policy usually performs poorly in the real environment. In order to improve the robustness of the controller against model inaccuracy and bridge the sim-to-real gap, we need to develop a robust control policy capable of handling various environments with different dynamics characteristics. To this end, we adopt dynamics randomization~\cite{sadeghi2016cad2rl,tobin2017domain} in our training strategy, in which dynamics parameters of the simulation environment are randomly sampled from an uniform distribution for each episode. 
The objective in Equation~(\ref{reward}) is then modified to maximize the expected reward across a distribution of dynamics characteristics $\rho(\mu)$:
\begin{equation}
\pi^* = \argmax_{\pi} \mathbb E_{\mu \sim \rho(\mu)}\mathbb E_{\tau \sim p(\tau |\pi,\mu)}[\sum \limits_{t=0}^{T-1}\gamma^t r_t],
\end{equation}
where $\mu$ represents the values of the dynamics parameters that are randomized during training. By training policies to adapt to variability in environment dynamics, the resulting policy will be more robust when transferred to the real world. 

\section{Numerical Experiment Results and Discussion}

We design a set of numerical experiments aiming to answer the following questions: 1) Can the learning process generate feasible control policies to control the exoskeleton to perform well-balanced squatting motions?  2) Will the learned control policies be robust enough under large random external perturbation? 3) Will the learned control policies be robust enough to sustain stable motions when subjected to uncertain human-exoskeleton interaction forces from a disabled human operator?

\subsection{Simulation and Reinforcement Learning Settings}
To demonstrate the effectiveness of our RL-based robust controller, we train the lower extremity exoskeleton to imitate a 4s reference squatting motion that is manually created based on human squatting motion. All 8 DoFs including on each side 1 DoF on the hip, 1 DoF on the knee, and 2 DoFs on the ankle are actuated to generate squatting motion. The open source library DART~\cite{Lee2018dynamic} is utilized to simulate the exoskeleton dynamics. The GRFs are computed by a Dantzig LCP (linear complementary problem) solver~\cite{baraff1994fast}. We utilize PyTorch~\cite{NEURIPS2019_9015} to implement the neural network and the PPO method for the learning process. The networks are initialized by the Xavier uniform method~\cite{glorot2010understanding}. We use a desktop computer with an Intel® Xeon(R) CPU E5-1660 v3 @ 3.00GHz × 16 to generate samples in parallel during training. Totally about 20 million samples are collected in each simulation. The policy and value networks are updated at a learning rate of $10^{-4}$, which is linearly decreased to 0 when 20 million samples are collected. PPO algorithm is robust in that hyperparameter initialization is a bit more forgiving and it can handle a wide variety of RL tasks. We do not deliberately tune hyperparameters and just use the common setting as in the literature. Hyper-parameters settings for training using PPO are shown in Table~\ref{Trainingsettings}. 

To verify the robustness of the trained controller, we test the control policies in out-of-distribution simulated environments, where the dynamic parameters of the exoskeleton are sampled randomly from a larger range of values than those during training. Table~\ref{Randomization} shows the dynamics parameters details of the exoskeleton and their range during training and testing. Note that the observation latency denotes the observation time delay in the real physical system due to sensor noise and time delay during information transfer. Considering the observation latency improves the reality of the simulations and further increases the difficulty for policy training.  The simulation frequency (time step for the environment simulation) and control policy output frequency are set to $900$Hz and $30$Hz, respectively. According to the PD torque Equation~(\ref{PD control}), the parameters about the  proportional gain $k_p$ and  differential gain $k_v$ are set to 900 and 40 respectively. The differential gain $k_v$ is chosen to be sufficiently high to prevent unwanted oscillation on the exoskeleton robot. From our experience, the control performance is robust against the variance of gains to a certain extent. For instance, increasing or decreasing the position gain $k_p$ to 1200 or 800 does not noticeably change the control performance. The reward weight settings are: $w^p= 0.8, w^{cop} = 0.8, w^v = 0.1, w^{ee} = 0.7, w^{com} = 0.4, w^{root} = 0.7, w^{torque} =0.1$.  The torque limit for each joint is set to 100$Nm$ in the simulation. 

\begin{table}[h]
\renewcommand\arraystretch{1.4}
	\caption{Hyper-parameters settings for training}
	\label{Trainingsettings}
	\begin{center}
		\begin{tabular}{ c c c c }
			\hline
			Parameters& Value &Parameters& Value\\
			\hline
			Discount factor   &    0.99 	&epochs  & 10\\
			Policy Adam learning rate &  $10^{-4}$  &clip threshold&  0.2\\
			batch size &128 &memory buffer &2048\\
			\hline
		\end{tabular}
	\end{center}
\end{table}

\begin{table}[h]
\renewcommand\arraystretch{1.4}
	\caption{Dynamic parameters and their respective range of values used during training and testing. A larger range of values are used during testing to evaluate the generalization ability of control policies in dynamics uncertainties.}
	\label{Randomization}
	\begin{center}
		\begin{tabular}{ c c c c }
			\hline
			Dynamic Parameters& Training Range & Testing Range\\
			\hline
			Friction coefficient   &    [0.9,1.6]*default value  &  [0.7,2.0]*default value\\
			Mass&    [0.8,1.2]*default value  & [0.7,1.5]*default value  \\
			Motor strength &[0.8,1.2]*default value  & [0.7,1.3]*default value \\
			Observation latency    &[0,0.04]s & [0,0.06]s\\
			Inertial&   [0.5,1.5]*default value  & [0.4,1.6]*default value\\
			Center of Mass& [0.9,1.2]*default value  & [0.8,1.3]*default value \\
			\hline
		\end{tabular}
	\end{center}
\end{table}

\subsection{Learned Squatting Skill}

\subsubsection{Case 1--feasibility demonstration}
\begin{figure}[h!]
\begin{center}
\includegraphics[width=0.98\columnwidth]{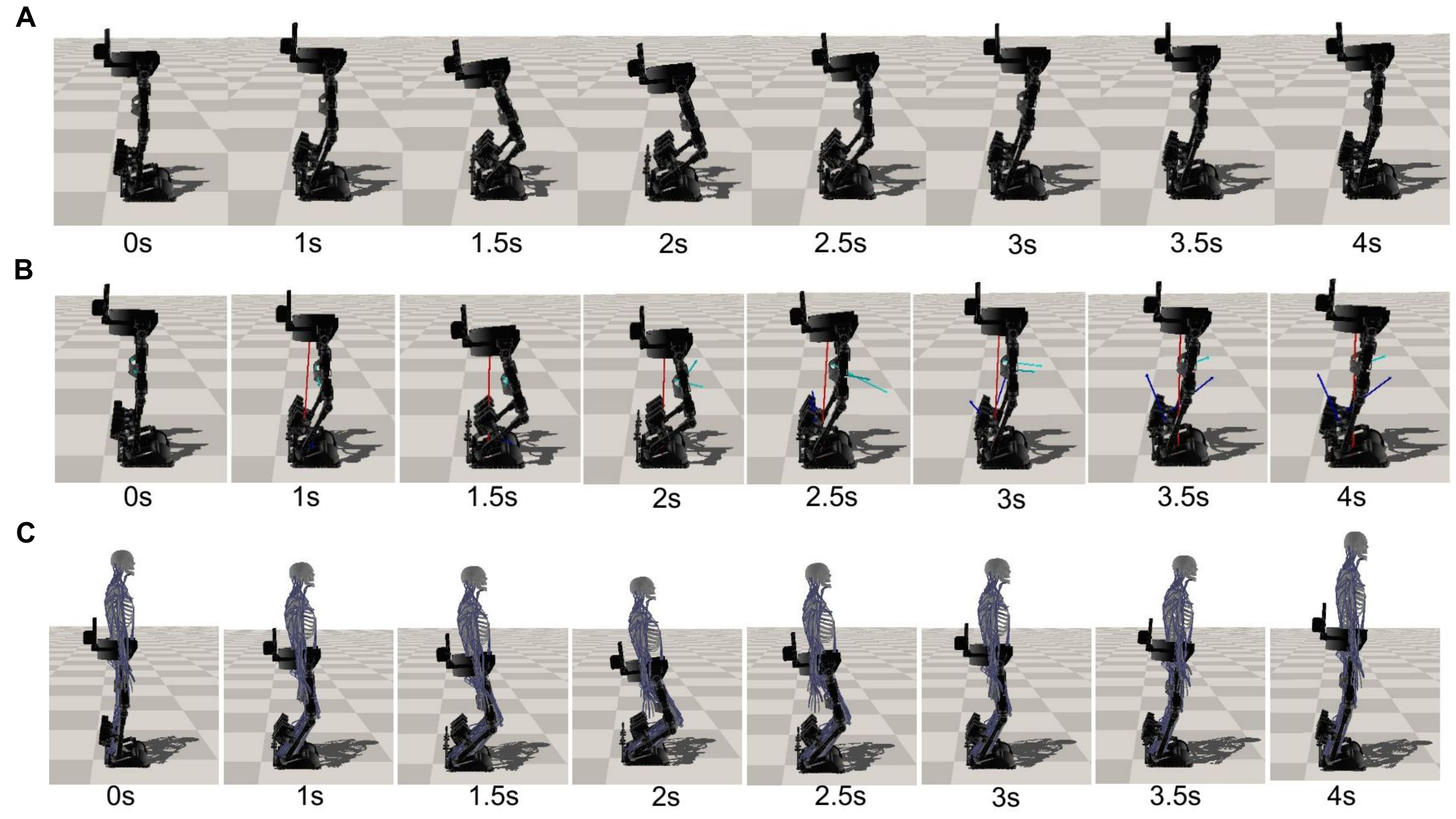}
\end{center}
\caption{Snapshots of the squatting motion of the exoskeleton. This RL-based controller enables the lower extremity exoskeleton to perform the squatting skill under a broad range of external perturbations. \textbf{(A)} performing the squatting skill without perturbation forces. \textbf{(B)} performing the squatting skill with large random perturbation forces. (Red, cyan and blue arrows show the random perturbation force applied on the hip, femur and tibia respectively.) \textbf{(C)} performing the squatting skill with human interaction.}
\label{fig:squatting_motion}
\end{figure}

\begin{figure}[!h]
	\begin{center}
	\includegraphics[width=0.9\columnwidth]{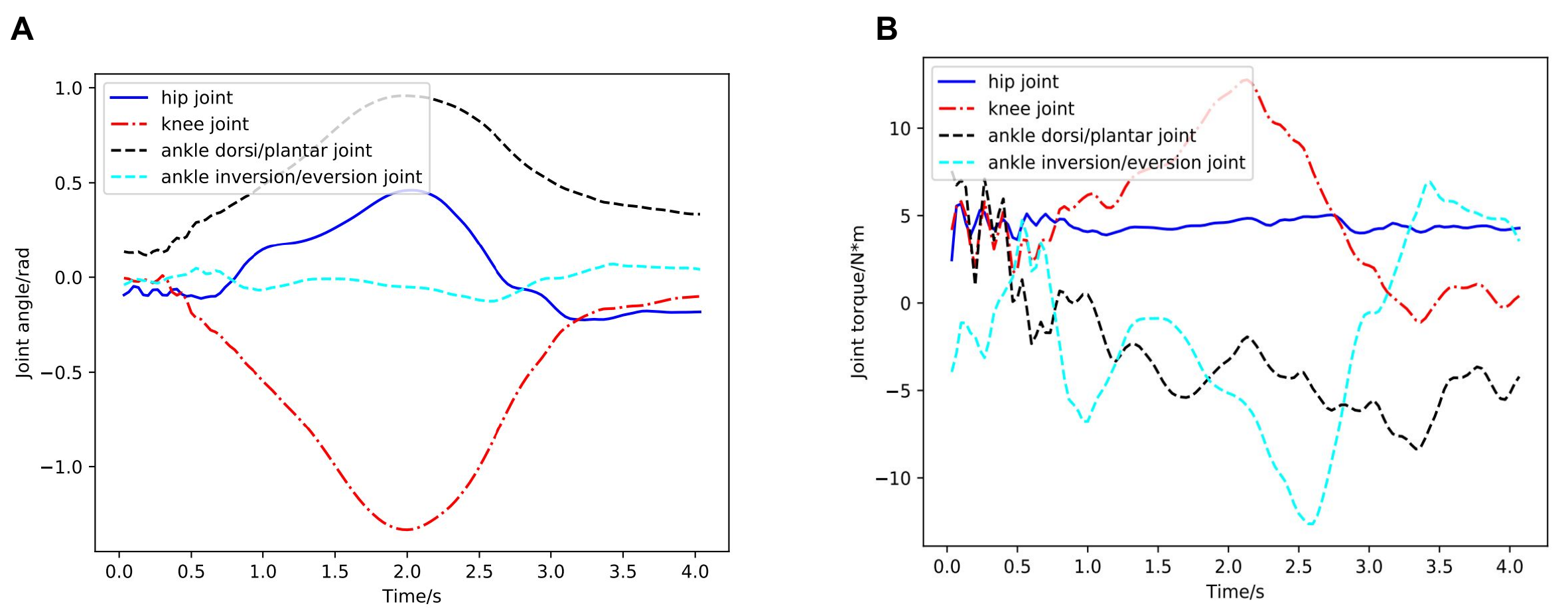}
    \end{center}
    \caption{Case1: joint behaviors in the first squatting cycle. \textbf{(A)} hip, knee, ankle dorsiflexion/plantarflexion, ankle inversion/eversion joint angles with respect to time during the first squatting cycle. \textbf{(B)} corresponding joint torques during the first squatting cycle.}
	\label{Case1_joint_angle_torque}
\end{figure}

In the first case, a squatting motion controller is learned from the 4s reference squatting motion without considering external perturbation. It is worth noting that, compared with the training, we use the larger-range dynamics randomization of the exoskeleton model to demonstrate the generalization ability of our learned controller (as shown in Table~\ref{Randomization}). A series of snapshots of the squatting behavior of the lower extremity exoskeleton under the learned control policy are shown in Figure~\ref{fig:squatting_motion}A. The lower extremity exoskeleton can perform the squatting and stand-up cycle with a nearly symmetric motion. Figure~\ref{Case1_joint_angle_torque}A displays the hip flexion/extension, knee flexion/extension, ankle dorsiflexion/plantarflexion and ankle inversion/eversion joint angles of the left leg in the first squatting cycle. Joint torques are depicted in Figure~\ref{Case1_joint_angle_torque}B and the peak torque at the knee joint for the squatting is around $13.5Nm$. Due to the weights of the waist support structure and the battery (around $2.2kg$) mounted on the back of it, the hip joint consistently produces a positive torque to prevent the waist structure from rotating downwards due to gravity. 

\begin{figure}[!h]
	\begin{center}
	\includegraphics[width=0.9\columnwidth,height=0.4\columnwidth]{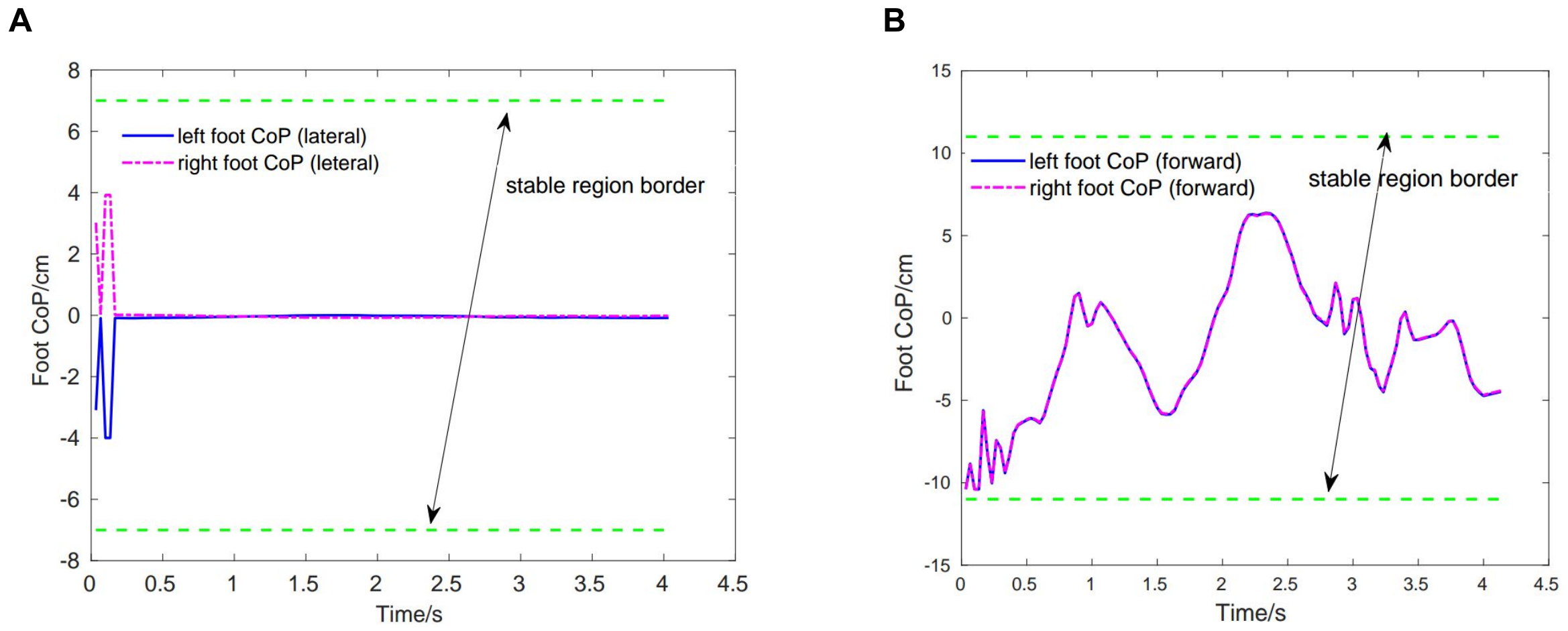}
    \end{center}
    \caption{Case1: foot CoP trajectories during the first squatting cycle. Green dotted line depicts the stable region border of CoP. \textbf{(A)} left foot CoP and right foot CoP trajectories (lateral direction). \textbf{(B)} left foot CoP and right foot CoP trajectories (forward direction).}
	\label{Case1_cop}
\end{figure}

Figure~\ref{Case1_cop} show the foot CoP trajectories of both left and right feet in the lateral and forward directions, which are calculated in real-time using the ground contact force information. As it can be seen, the exoskeleton controller can keep the foot CoP well inside the stable region for both lateral and forward directions in a complete squatting cycle. Noted at the beginning the CoP is close to the back edge (due to its initial state) but gradually it is bought to near the center. This indicates that the exoskeleton controller is able to recognize the current state of CoP and capable of bring it to a more stable state. And the balance in the lateral direction is better than that in the forward direction due to the symmetric nature of the squatting motion. The right foot CoP trajectories have very similar patterns with those of the left foot CoP. 

\begin{figure}[!h]
	\begin{center}
	\includegraphics[width=0.5\columnwidth]{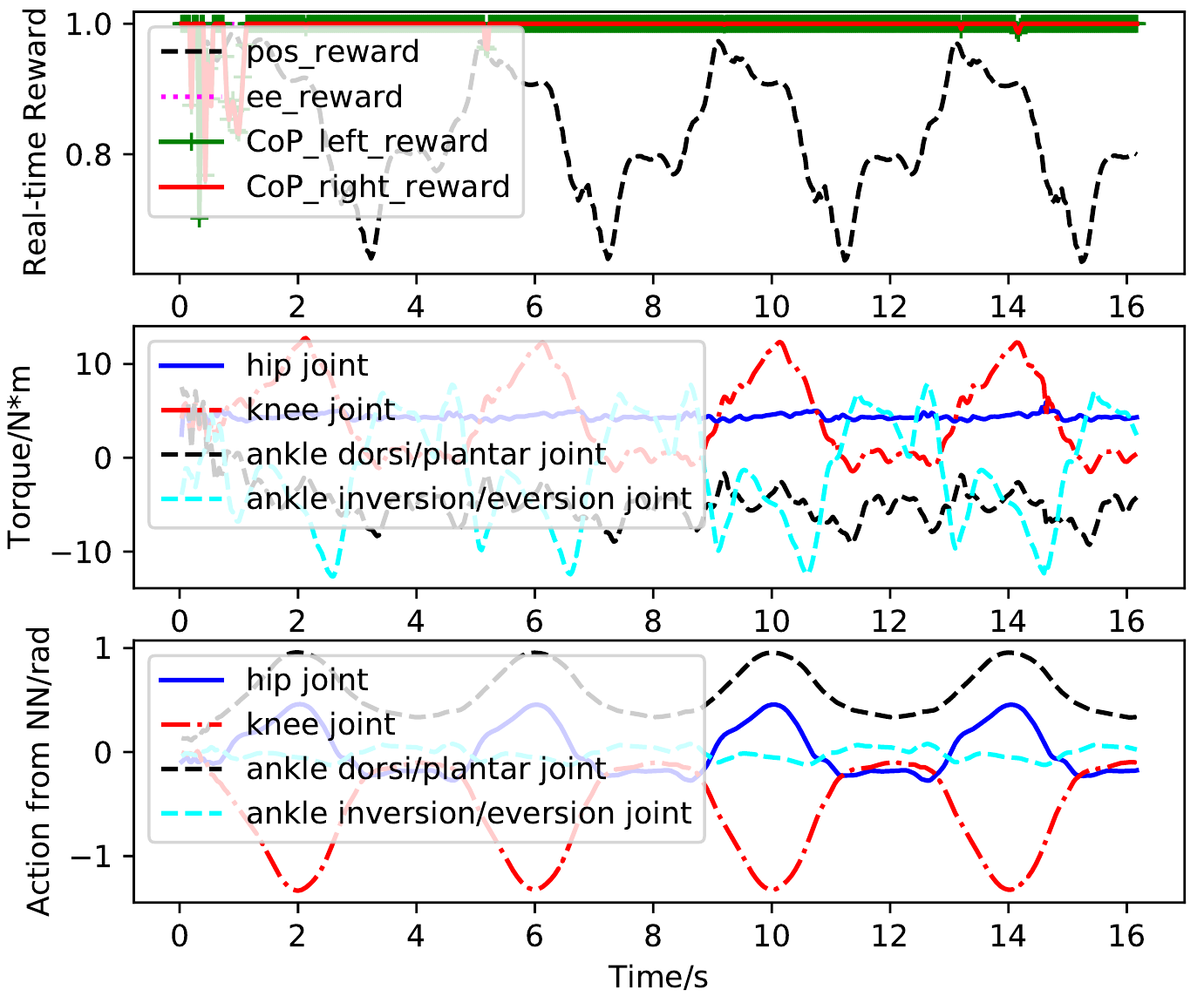}
	 \end{center}
	\caption{Case1: performance of the RL based controller without external perturbation during multiple squatting cycles. The first figure demonstrates the foot CoP reward (green and red lines) , end-effector tracking reward (magenta line) and joint position tracking reward (black line) calculated according to Equation~(\ref{pos_reward})-(\ref{cop_reward}). The bottom figure depicts the actions for the hip, knee ankle joint predicted from the neural network. }
	\label{Case1_tracking}
\end{figure}

Figure~\ref{Case1_tracking} presents the performance of the learned controller while performing multiple squatting cycles. We can clearly observe the high CoP rewards, indicating good system balance when the exoskeleton performs the squatting motion. The relatively high joint position tracking and end-effector tracking rewards illustrate strong tracking performance of the control system. The second figure shows the torques for the hip, knee, and ankle joints. The last figure demonstrates the predicted actions (PD target positions) for these joints. It is clear from these plots that the actions predicted from the policy network are smooth and exhibit clear cyclic patterns. 

\subsubsection{Case 2--robustness against random perturbation forces}

In the second case, we aim to verify the robustness of the controller under random external perturbation forces. From our tests, the learned control policy from case 1, trained without any perturbation forces, could perform well with random perturbation forces up to $100N$. To further improve the robustness of our controller to handle greater perturbation forces, we now introduce random perturbation forces during the training process of the learning controller. The perturbation forces are applied to three parts of the exoskeleton: hip, femur, and tibia. For femur and tibia, the magnitude of forces are randomly sampled in the range $(0, 100)N$ and no restriction is set for the direction. For the hip, we randomly sample the magnitude of force in $(0,200)N$ but restricted the direction to $0\sim 20$ degrees from the vertical direction assuming no large lateral pushing perturbation. 

\begin{figure}[!h]
	\begin{center}
	\includegraphics[width=0.9\columnwidth]{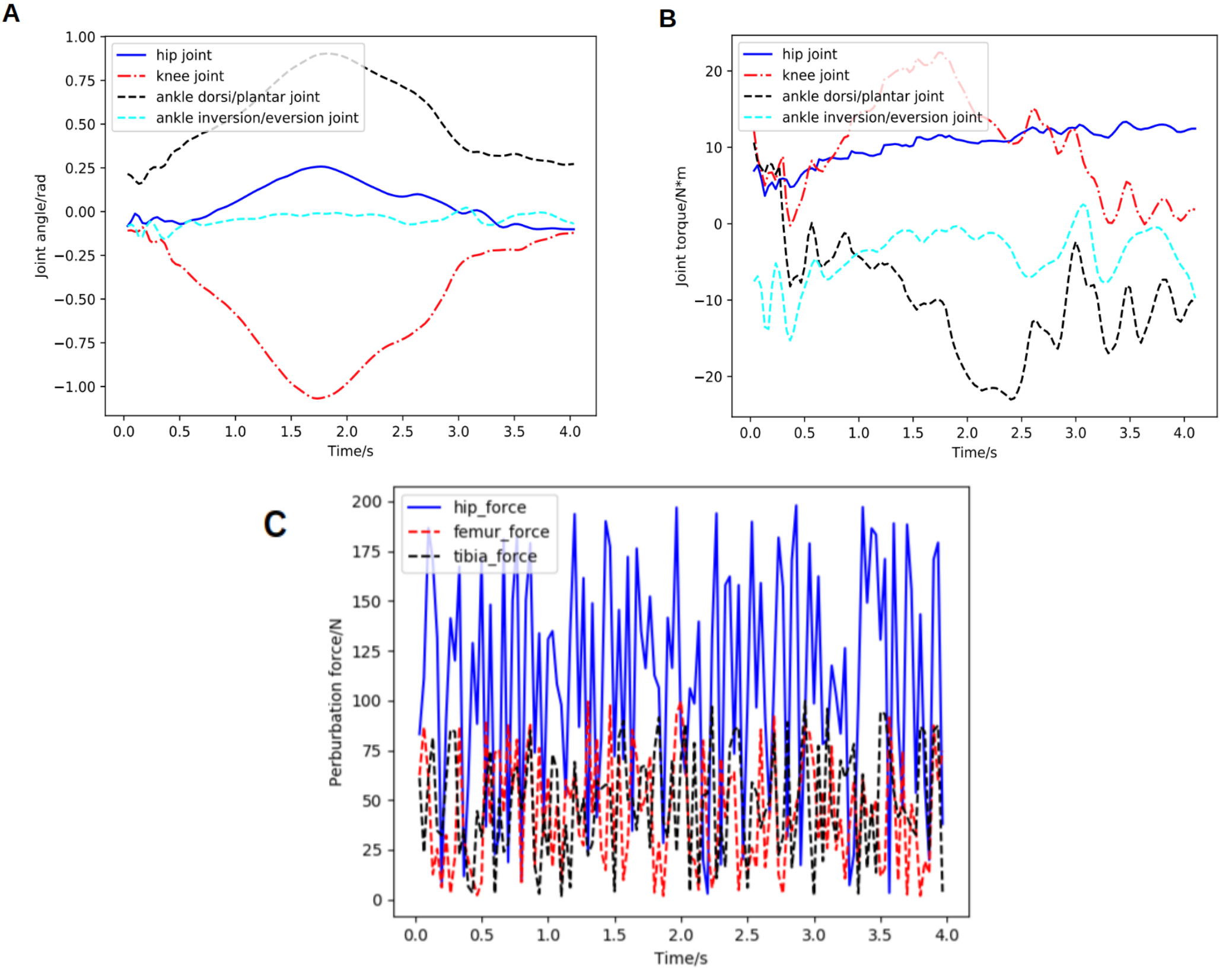}
    \end{center}
    \caption{Case2: joint behaviors under random perturbation forces in the first squatting cycle. \textbf{(A)} hip, knee, ankle dorsiflexion/plantarflexion, ankle inversion/eversion joint angles with respect to time in the first squatting cycle. \textbf{(B)} corresponding joint torques in the first squatting cycle.  \textbf{(C)} magnitudes of the random perturbation forces applied on the three parts of the exoskeleton: hip, femur and tibia during testing.}
	\label{Case2_joint_angle_torque}
\end{figure}


\begin{figure}[!h]
	\begin{center}
	\includegraphics[width=0.96\columnwidth,height=0.6\columnwidth]{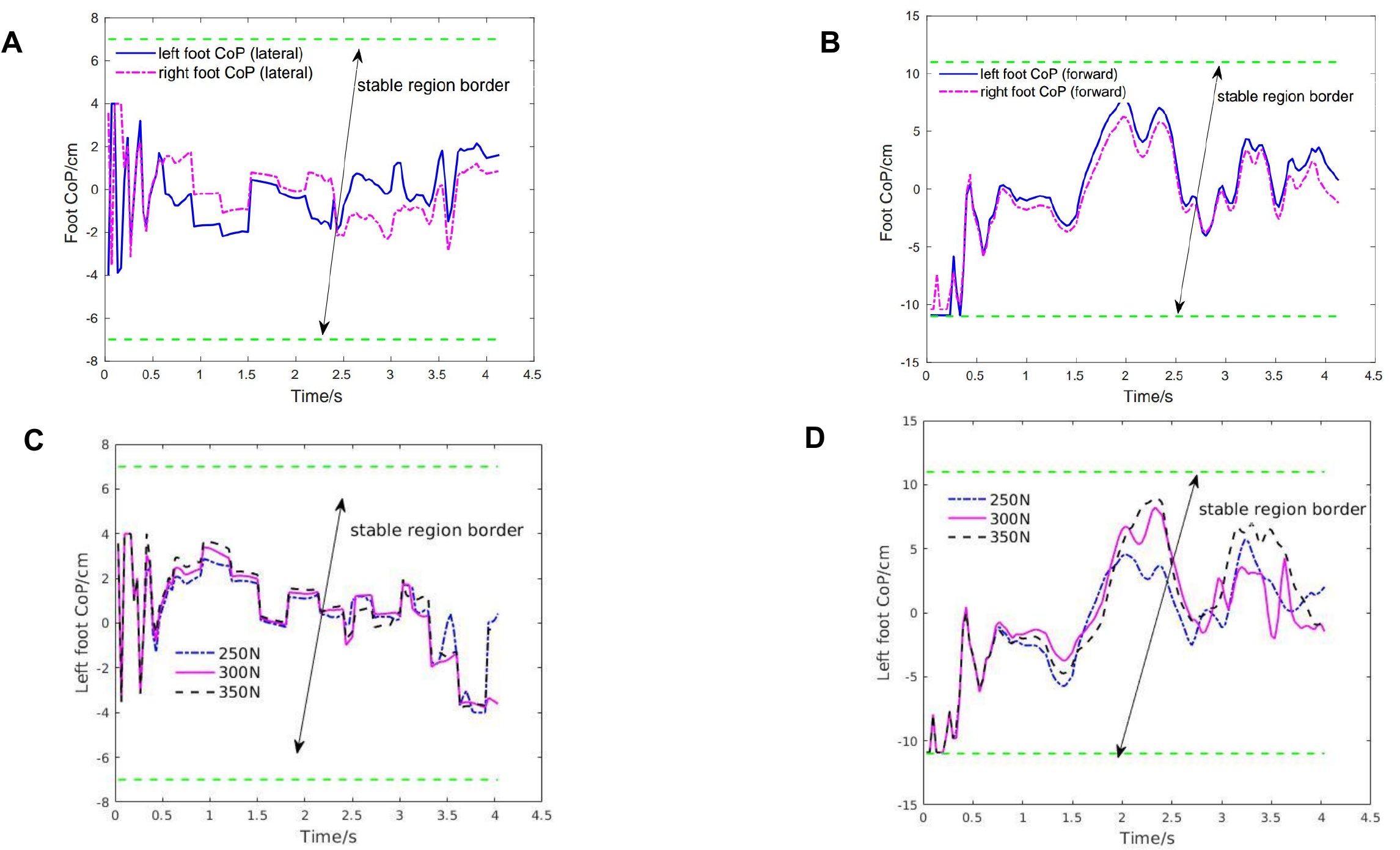}
    \end{center}
    \caption{Case2: foot CoP trajectories under random external perturbations in the first squatting cycle and robustness test on the foot CoP stability under 75\% greater perturbation forces compared with the training setting. \textbf{(A)} left and right foot CoP trajectories (lateral direction). \textbf{(B)} left and right foot CoP trajectories (forward direction). \textbf{(C)} left foot CoP trajectory (lateral direction) under 75\% greater perturbation forces compared with the training setting. \textbf{(D)} left foot CoP trajectory (forward direction) under greater perturbation forces compared with the training setting.}
	\label{Case2_cop}
\end{figure}


Figure~\ref{fig:squatting_motion}B shows a series of snapshots of the lower extremity exoskeleton behavior with the newly trained controller when tested with random, large perturbation forces during squatting motion. Joint angles and torques at the hip, knee and ankle joint of the left leg in the first squatting cycle are shown in Figure~\ref{Case2_joint_angle_torque}. We can clearly observe that the motion is still relative smooth and the torques calculated from Equation~(\ref{PD control}) have more ripples in response to the random perturbation forces. Figure~\ref{Case2_joint_angle_torque}C shows randomly varying perturbation forces applied on the hip, femur and tibia.
Compared to the joint angles and torques without external perturbation (Case 1, Figure~\ref{Case1_joint_angle_torque}), the joint torques under random perturbation forces are almost doubled but the joint angles are relatively close. Moreover, under the random perturbations that could influence the balance in the frontal plane, the learned controller can still keep balance and stability with the actuation of the ankle inversion/eversion joint.

\begin{figure}[!h]
	\begin{center}
	\includegraphics[width=0.5\columnwidth,height=0.4\columnwidth]{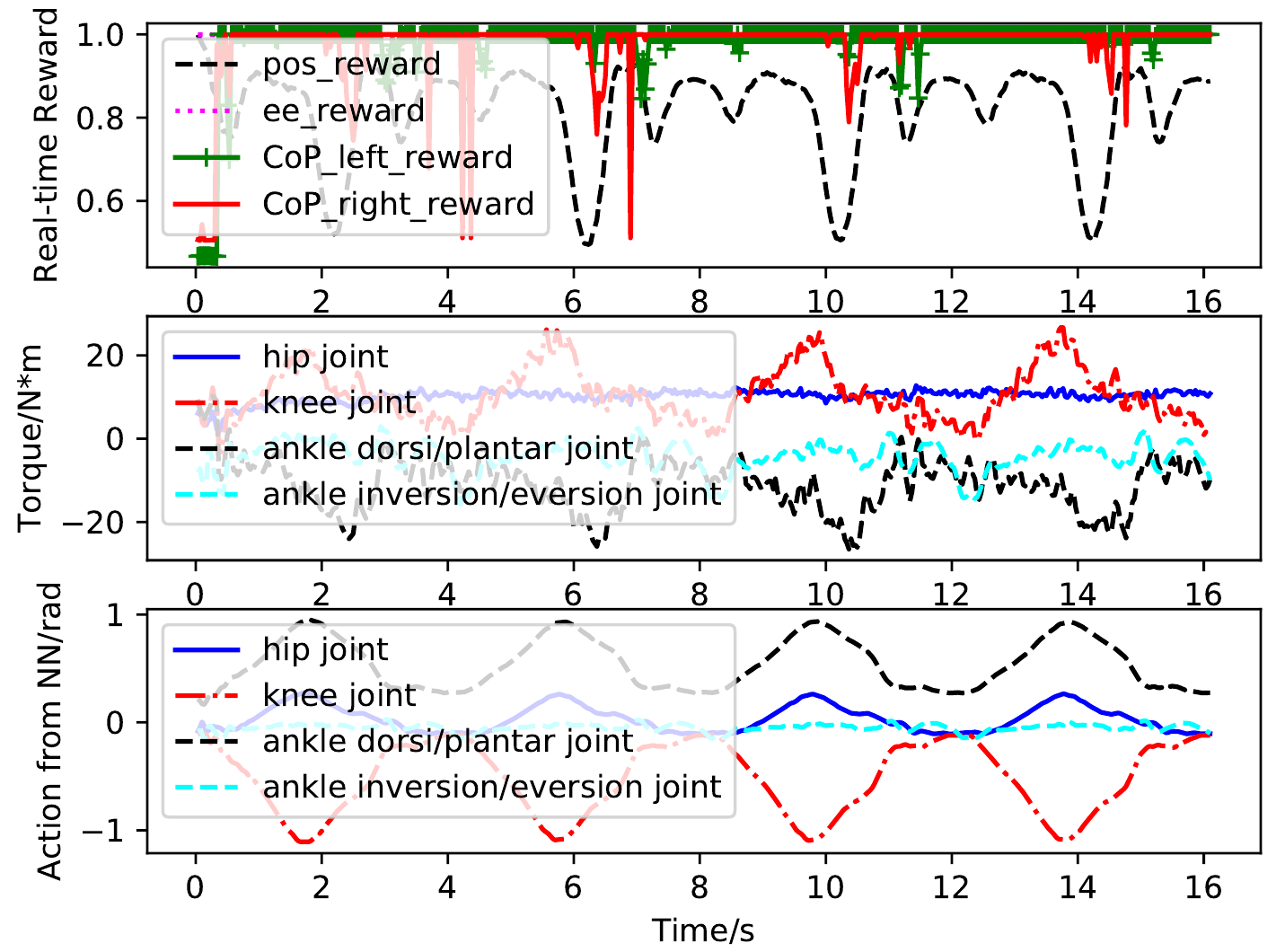}
	 \end{center}
	\caption{Case2: real-time performance of the reinforcement-learning based controller with large, random external perturbation during multiple squatting cycles. The first figure demonstrates the real-time foot CoP reward (green and red lines), end-effector tracking reward (magenta line) and joint position tracking reward (black line) calculated according to Equation~(\ref{pos_reward})-(\ref{cop_reward}). The bottom figure depicts the actions for the hip, knee ankle joint predicted from the neural network.}
	\label{Case2_tracking}
\end{figure}

Figure~\ref{Case2_cop} shows the CoP trajectories of the left and right feet in both lateral and forward directions in the first squatting cycle. As shown in this figure, the foot CoP trajectories have oscillation under the random perturbation forces compared with Case1. But the robot can still keep the CoP inside the safe region of the support foot to guarantee stability and balance. It further validates that this controller enables the robot to perform squatting motion with strong stability and robustness. To further demonstrate the robustness of the learning controller, we increase the random perturbation forces up to $75\%$ greater than the training perturbation (e.g. up to $350N$ for the hip force) and found that the exoskeleton can still perform the squatting motion without losing its balance as illustrated in Figure~\ref{Case2_cop}C-D about the left foot CoP stability. Real-time tracking results of the exoskeleton with multiple squatting cycles are shown in Figure~\ref{Case2_tracking}. From Figure~\ref{Case2_tracking}, both the foot CoP reward and the end-effector reward remain high under the random perturbation forces. The torques are greater than that without external perturbation forces, the peak torque for the knee joint is close to $40Nm$. The actions predicted from the policy network are also smooth and exhibit cyclic patterns.

\subsubsection{Case3--robustness under human-exoskeleton interaction}
\begin{figure}[!h]
	\begin{center}
	\includegraphics[width=0.9\columnwidth]{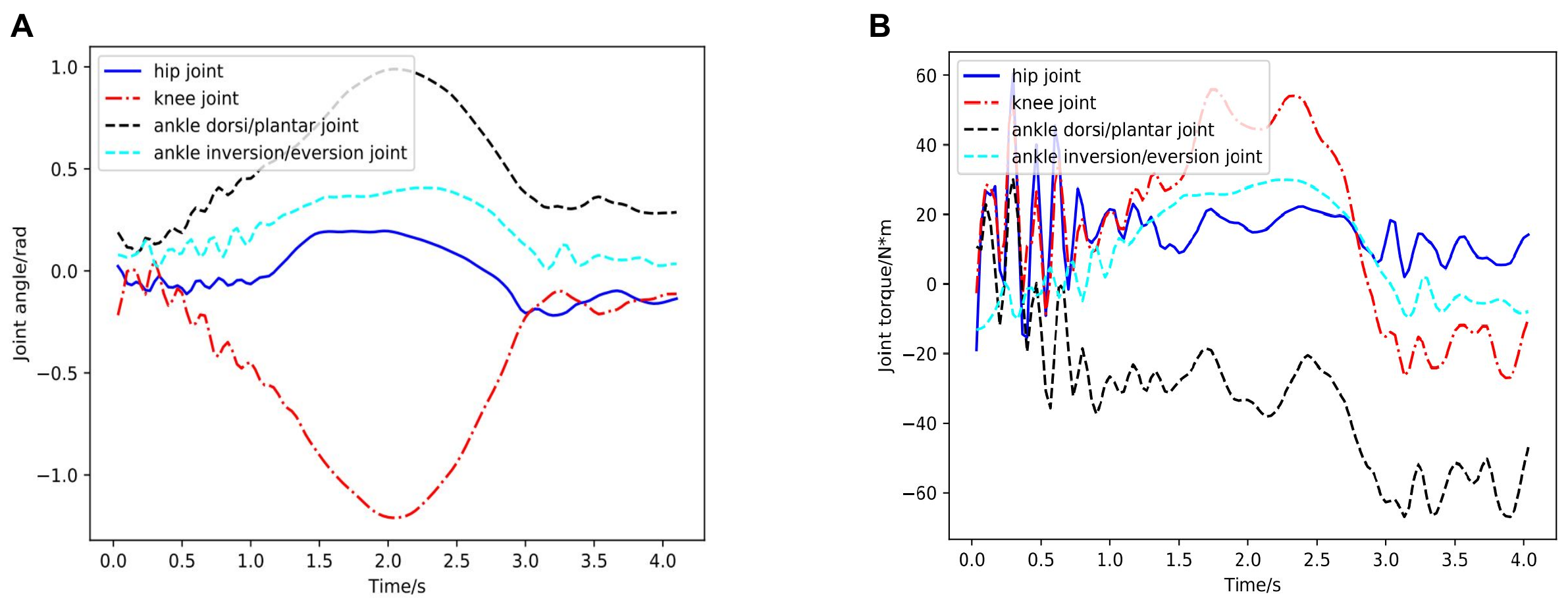}
    \end{center}
    \caption{Case3: joint behaviors under human-exoskeleton interactions during the first squatting cycle. \textbf{(A)} hip, knee, ankle dorsiflexion/plantarflexion, ankle inversion/eversion joint angles in the first squatting cycle. \textbf{(B)} corresponding joint torques in the first squatting cycle. \textbf{(C)} human-exoskeleton interaction (strap) forces during multiple squatting cycles. hip$\_$force, femur$\_$force and tibia$\_$force in the figure represent the interaction forces between the human and the strap on the exoskeleton respectively.}
	\label{Case3_joint_angle_torque_human_interaction}
\end{figure}

\begin{figure}[!h]

	\begin{center}
	\includegraphics[width=0.9\columnwidth,height=0.36\columnwidth]{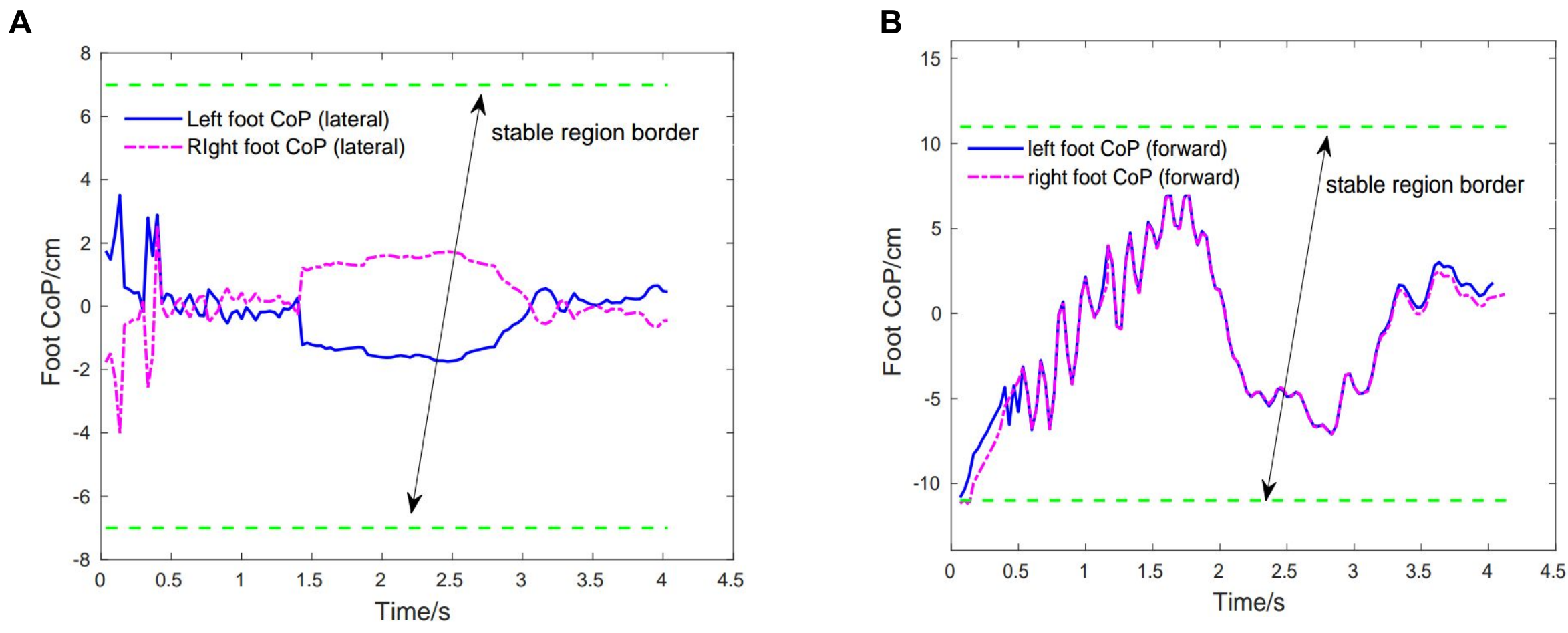}
    \end{center}
    \caption{Case3: foot CoP trajectories under human-exoskeleton interactions in the first squatting cycle. The stable region border is marked with green dotted lines. \textbf{(A)} left foot CoP and right foot CoP trajectories (lateral direction). \textbf{(B)} left foot CoP and right foot CoP trajectories (forward direction).}
	\label{Case3_cop}
\end{figure}

\begin{figure}[!h]
	\centering
	\includegraphics[width=0.95\columnwidth]{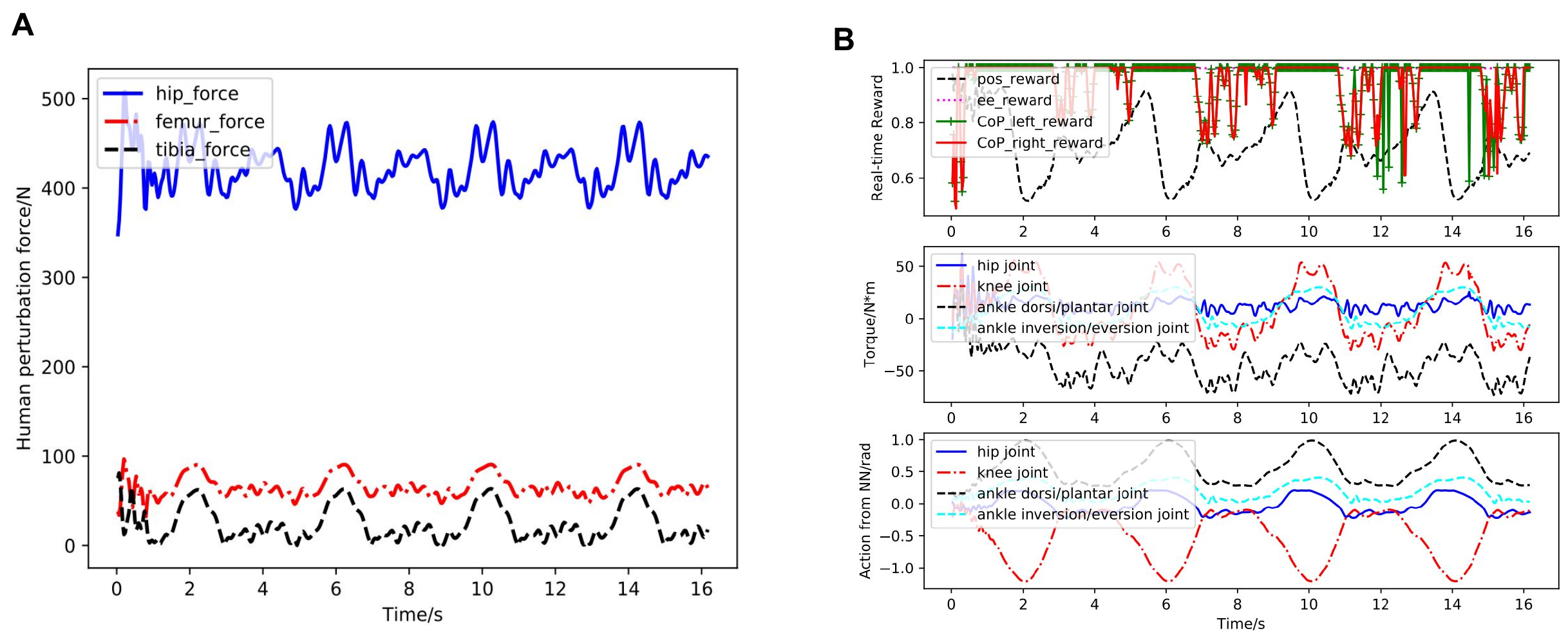}
	\caption{Case3: performance of the RL based controller under human-exoskeleton interactions during more squatting cycles. \textbf{(A)} human-exoskeleton interaction (strap) forces during multiple squatting cycles. hip$\_$force, femur$\_$force and tibia$\_$force in the figure represent the interaction forces between the human and the strap on the exoskeleton respectively. \textbf{(B)} performance of the RL based controller. The first figure demonstrates the foot CoP reward (green and red lines) , end-effector reward (magenta line) and joint position tracking reward (black line) calculated according to Equation~(\ref{pos_reward})-(\ref{cop_reward}). The bottom figure depicts the actions for the hip, knee ankle joint predicted from the neural network.}
	\label{Case3_tracking}
\end{figure}

In the third case, the human musculoskeletal model is integrated with the exoskeleton to simulate more realistic perturbation forces. As described in the section~\ref{human_interaction}, the springs at the strap locations can generate the varying interaction forces between the human and exoskeleton robot during the motion, which are applied on both human and exoskeleton. We first train the network with the integrated human-exoskeleton model to account for the interaction forces. Here we do not consider the active muscle contraction of the human operator or actuation torques on the human joints, considering the operator could be a patient suffering from spinal cord injury or stroke, with very limited or no control of his or her own body. Nonetheless, the passive muscle forces as described in \cite{lee2019scalable} during movement are incorporated. The squatting skill learned by the exoskeleton and performance of the motion controller are shown in Figure~\ref{fig:squatting_motion}C and Figure~\ref{Case3_joint_angle_torque_human_interaction}-\ref{Case3_tracking}.

As shown in Figure~\ref{fig:squatting_motion}C, the rehabilitation exoskeleton is able to assist the human to perform the squatting motion without external assistance. Angles and torques at the hip, knee and ankle joint of the left leg in the first cycle are shown in Figure~\ref{Case3_joint_angle_torque_human_interaction}A-B. The torques at the hip, knee and ankle joint are greater than those without the human interaction while still below the maximum torques. Figure~\ref{Case3_cop} shows the CoP trajectories of the left foot and right foot in the lateral and forward directions under the predicted human-skeleton interaction forces (as shown in Figure~\ref{Case3_tracking}A).
From Figure~\ref{Case3_cop}A, there exists a symmetric pattern between the left and right foot CoP trajectories in the lateral direction, which indicates that the squatting motion is well balanced in the frontal plane. Real-time tracking results of the controlled exoskeleton under the human-exoskeleton interactions with more squatting cycles are shown in Figure~\ref{Case3_tracking}. As shown in Figure~\ref{Case3_tracking}B, the high CoP reward in the first figure indicates that the proposed control system has strong stability and robustness to the human interaction forces. The peak torque for all joints are less than $70Nm$. The largest peak torque happens at the ankle dorsi/plantar flexion joint ($68Nm$), which is much smaller than its 160$Nm$ capacity~\cite{nunez20172}. The actions predicted from the policy network, shown in the last figure of Figure~\ref{Case3_tracking},  are still smooth and cyclic in general. 

\begin{figure}[!h]
	\centering
	\includegraphics[width=0.95\columnwidth]{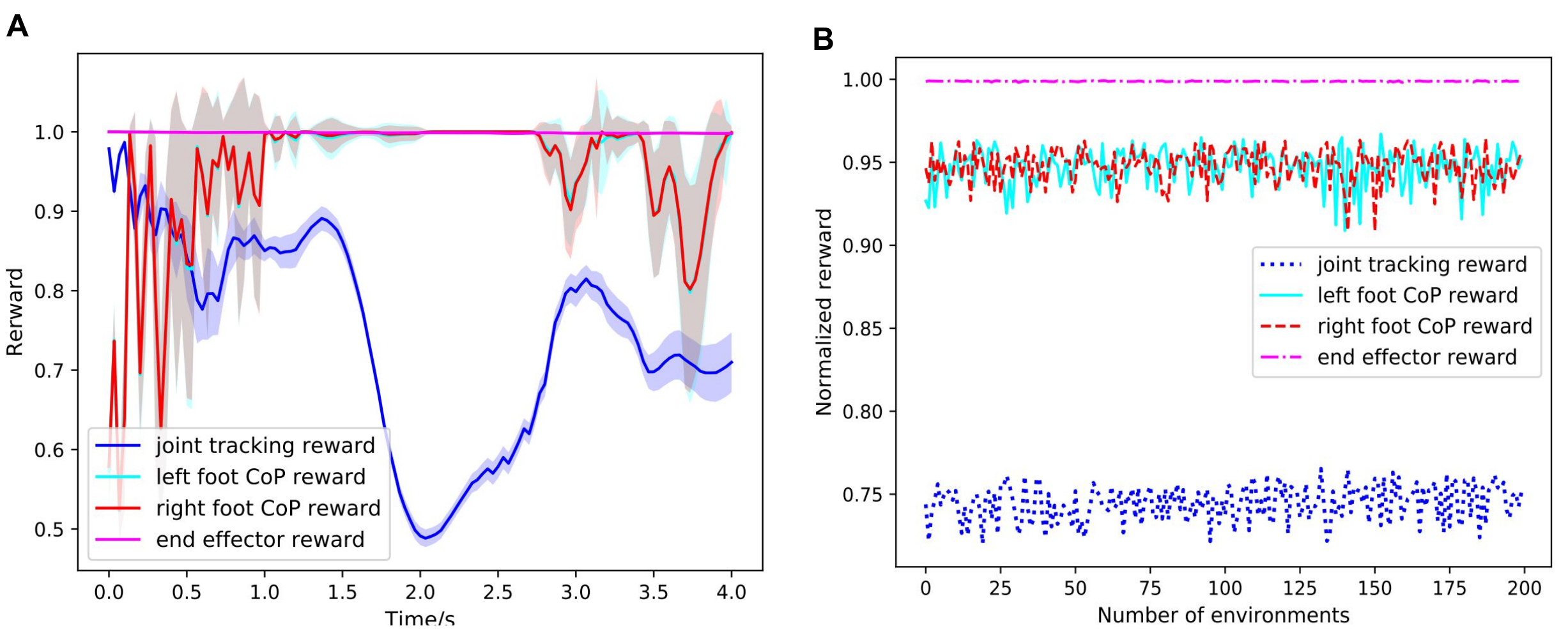}
	\caption{Performance of learned controller under human-exoskeleton interactions in 200 simulated environments with different dynamics. The dynamic model of the exoskeleton is randomly initialized. \textbf{(A)} Rewards statistics (curve: mean; shade: standard deviation) with respect to time for 200 simulated environments. \textbf{(B)} Average reward of a complete squatting cycle for each simulated environment.
	}
	\label{robustness_control_policies}
\end{figure}

To further validate the learned controller's ability to cope with unfamiliar dynamics of the exoskeleton, we test the learned controller in 200 out-of-distribution simulated environments, where the dynamics parameters are sampled from a larger range of values than those used during training (as shown in Table~\ref{Randomization}). Figure~\ref{robustness_control_policies} visualizes the performance of the learned controller in 200 simulated environments with different dynamics. Figure~\ref{robustness_control_policies}A depicts the rewards statistics (mean and standard deviation) with respect to time calculated from Equation~(\ref{pos_reward})-(\ref{cop_reward}) under 200 simulated environments. The end-effector reward indicating the foot tracking performance consistently maintains a high value, revealing the exoskeleton robot has no falling condition in 200 simulated environments with unfamiliar dynamics and it can stand on the ground with stationary feet when performing the complete squatting motion. The joint position tracking and foot CoP also achieve a high reward with less variance under more diverse dynamics of the exoskeleton robot. Figure~\ref{robustness_control_policies}B shows the average reward of a complete squatting cycle for each simulated environment. These results suggest that the learned controller is able to effortlessly generalize to environments that differ from those encountered during training and achieve good control performance under very diverse dynamics. This also suggests that the learned controller will likely be robust when transferred to the real exoskeleton. 

\section{Discussion}

Through these designed numerical tests, we verified that our learning framework can produce effective neural network-based control policies for the lower extremity exoskeleton to perform well-balanced squatting motions. By incorporate adverse perturbations in training, the learned control policies are robust against large random external perturbations during testing. And it can sustain stable motions when subjected to uncertain human-exoskeleton interaction forces from a disabled human user. From all numerical tests performed, the effectiveness and robustness of the learned balance controller are demonstrated by its capability to maintain CoP inside the foot stable region along both lateral and forward directions.

In this study, we evaluated the controllers in a specific case for which the human musculoskeletal model has only passive muscle response (i.e. without active muscle contraction). In reality, the human-exoskeleton interaction forces might vary substantially for different users with different weights and levels of disability. One may assume that a user with good body or muscle control tends to minimize the interaction forces on the straps. On the other hand, a passive human musculoskeletal model tends to generate larger interaction forces. Therefore, using a passive musculoskeletal model can be considered a more difficult task for the exoskeleton. 
Further investigations with an active human musculoskeletal model are possible but will likely to need additional information on the health condition of the user. 

Through dynamics randomization, our learned controller is robust against modeling inaccuracy and differences between the training and testing environments~\cite{tan2018sim}. Nonetheless, it is still beneficial to validate and further improve the dynamic exoskeleton model. This can be done through a model identification process, for which we can conveniently rely on the real-time measurement of GRFs and derived CoP information from the foot force sensors during programmed motions. Experiments can be conducted to correlate the CoP position with the exoskeleton posture and the CoP movement with motion. Stable squatting motions of different velocities can be used to record the dynamic responses of the exoskeleton and the collected data then can be used to further identify or optimize the model’s parameters (such as inertia properties and joint friction coefficients). 

By incorporating motion imitation into the learning process, the proposed robust control framework has the potential to learn a diverse array of human behaviors without the need to design skill-specific reward functions. Common rehabilitation motions such as sit-to-stand, walking on flat or inclined ground can be learning by feeding proper target motions. Due to the nature of imitation learning and CoP based balance control, we foresee minimal changes to the learning framework with the exception of crafting different target motions for imitation. The learning process will automatically create specific controllers that can produce physically feasible and stable target motions (even when the target motion is coarsely generated and may not be physically feasible). 

Transferring or adapting RL control policies trained in the simulations to the real hardware remains a challenge in spite of some demonstrated successes. To bridge the so-called "sim-to-real" gap, we have adopted dynamics randomization during training that is often used to prepare for sim-to-real transfer~\cite{tan2018sim}. In a recent work by Exarchos et al.~\cite{exarchos2020policy}, it is shown kinematic domain randomization can also be effective for policy transfer. Additionally, efficient adaptation techniques such as latent space ~\cite{8968053,clavera2018learning} or meta-learning ~\cite{yu2020learning} can also be applied to further improve the performance of pre-trained policies in the real environment. We plan to construct an adaptation strategy to realize the sim-to-real control policy transfer for the lower extremity exoskeleton robot in the near future. 

\section{Conclusion}
In this work, we have presented a robust, RL-based controller for exoskeleton squatting assistance with human interaction. A relatively lightweight lower extremity exoskeleton is presented and used to build a human-exoskeleton interaction model in the simulation environment for learning the robust controller. The exoskeleton foot CoP information collected from the force sensors is leveraged as a feedback for balance control and adversary perturbations and uncertain human interaction forces are used to train the controller. We have successfully demonstrated the lower extremity exoskeleton's capability to carry a human to perform squatting motions with a moderate torque requirement and provided evidence of its effectiveness and robustness. With the actuation of the ankle inversion/eversion joint, the learned controllers are also capable of maintaining the balance within the frontal plane under large perturbation or interaction forces. The success demonstrated in this study has implications for those seeking to apply reinforcement learning to control robots to imitate diverse human behaviors with strong balance and increased robustness even when subjected to the larger perturbations. 
In the near future, We plan to extend this framework to learn a family of controllers that can perform a variety of human skills like sit-to-stand, flat and inclined ground walking. Lastly, these learned controllers will be deployed to the hardware through a sim-to-real transfer method and experimental tests will be performed to validate the controllers' performance.

\bibliographystyle{unsrt}  
\bibliography{references}  

\end{document}